\documentclass[conference]{IEEEtran}
\IEEEoverridecommandlockouts
\usepackage{cite}
\usepackage{amsmath,amssymb,amsfonts}
\usepackage{algorithmic}
\usepackage{graphicx}
\usepackage{textcomp}
\usepackage{xcolor}
\def\BibTeX{{\rm B\kern-.05em{\sc i\kern-.025em b}\kern-.08em
    T\kern-.1667em\lower.7ex\hbox{E}\kern-.125emX}}
\begin{document}

\title{ConvexECG: Lightweight and Explainable Neural Networks for Personalized, Continuous Cardiac Monitoring\\
\thanks{$\dagger$ These authors contributed equally to this work. \\ 
This work was supported in part by the National Science Foundation (NSF) under Grant DMS-2134248 and the CAREER Award Grant CCF-2236829; the Office of Naval Research under Grant N00014-24-1-2164; the National Institutes of Health (NIH) under Grants K23, R01-HL162260, and R01-HL149134; and the American Heart Association (AHA) under the Career Development Award.
}
}

\author{\IEEEauthorblockN{Rayan Ansari$^\dagger$}
\IEEEauthorblockA{\textit{Department of Bioengineering} \\
\textit{Stanford University School of Engineering}\\
Stanford, USA \\
0000-0001-6153-7272}
\and
\IEEEauthorblockN{John Cao$^\dagger$}
\IEEEauthorblockA{\textit{Department of Electrical Engineering} \\
\textit{Stanford University School of Engineering}\\
Stanford, USA \\
0009-0009-6916-1295}
\and
\IEEEauthorblockN{Sabyasachi Bandyopadhyay}
\IEEEauthorblockA{\textit{Department of Medicine} \\
\textit{Stanford University School of Medicine}\\
Stanford, USA \\
0009-0003-4825-646X}
\and
\IEEEauthorblockN{Sanjiv M. Narayan}
\IEEEauthorblockA{\textit{Department of Cardiovascular Medicine} \\
\textit{Stanford University School of Medicine}\\
Stanford, USA \\
0000-0001-7552-5053}
\and
\IEEEauthorblockN{Albert J. Rogers}
\IEEEauthorblockA{\textit{Department of Cardiovascular Medicine} \\
\textit{Stanford University School of Medicine}\\
Stanford, USA \\
0000-0001-6585-534X}
\and
\IEEEauthorblockN{Mert Pilanci}
\IEEEauthorblockA{\textit{Department of Electrical Engineering} \\
\textit{Stanford University School of Engineering}\\
Stanford, USA \\
0000-0002-0870-9992}
}

\maketitle

\begin{abstract}
We present ConvexECG, an explainable and resource-efficient method for reconstructing six-lead electrocardiograms (ECG) from single-lead data, aimed at advancing personalized and continuous cardiac monitoring. ConvexECG leverages a convex reformulation of a two-layer ReLU neural network, enabling the potential for efficient training and deployment in resource constrained environments, while also having deterministic and explainable behavior. Using data from 25 patients, we demonstrate that ConvexECG achieves accuracy comparable to larger neural networks while significantly reducing computational overhead, highlighting its potential for real-time, low-resource monitoring applications.
\end{abstract}


\section{Introduction}

Continuous electrocardiogram (ECG) monitoring plays a critical role in diagnosing and managing cardiac conditions. Implantable cardiac monitors (ICM) and wearable devices (e.g., smartwatches) offer convenience, but their single-lead format limits more complex ECG analyses, such as arrhythmia localization, diagnosis of conduction abnormalities, QT interval measurement, and ischemia detection. Multi-lead ECG monitoring provides a more comprehensive view of the heart’s electrical activity, allowing clinicians to detect and localize abnormalities that may be visible in certain leads but not others, thus significantly expanding diagnostic utility. Accurately diagnosing these factors is essential for assessing the risk of critical adverse cardiac events such as out-of-hospital cardiac arrest and sudden cardiac death.

Therefore, to enhance the clinical utility of single-lead cardiac monitoring devices, much research has pursued the task of lead reconstruction from reduced lead sets \cite{tragardh_how_2006}. Various mathematical models and signal processing techniques have been employed to determine feasible reduced lead sets that preserve the diagnostic integrity of the complete monitoring setup. Approaches such as linear regression provide simplicity and explainability but lack the capacity to capture the nonlinear relationships between ECG leads \cite{nelwan_minimal_2000, nelwan_reconstruction_2004}. Meanwhile, deep learning models, though more expressive, are resource-intensive and opaque, making them unsuitable for point-of-care deployment and in providing clinical transparency \cite{wang_novel_2019, smith_reconstruction_2021, joo_twelve-lead_2023}.

To address these limitations, we propose ConvexECG, a convex formulation of two-layer rectified linear unit (ReLU) neural networks \cite{pilanci2020neural} for reconstructing 6-lead ECGs from single-lead data. 

Our hypothesis is that convex neural networks, by combining the universal approximation power of neural networks \cite{PALUZOHIDALGO202029} with the tractability and guaranteed global optimality of convex optimization, can maintain the diagnostic capability of reduced lead set ECG reconstruction while significantly improving computational efficiency and interpretability. This approach represents the first application of convex neural networks in ECG reconstruction, offering a unique set of advantages:

\begin{enumerate}
    \item Inherent explainability via its convex formulation.

    \item Lightweight, low-complexity without sacrificing expressiveness.

    \item Deterministic behavior with guaranteed convergence to a global optimum.

\end{enumerate}

The outline of the paper is as follows: Section \ref{prelimiaries} briefly introduces the theory of convex neural networks, Section \ref{problem} formalizes the single-lead ECG reconstruction problem, Section \ref{approach} presents our proposed method, Section \ref{experiments} demonstrates its application on real ECG data. Lastly, Section \ref{conclusion} concludes the paper with important takeaways and potential future research directions. 

\section{Preliminaries on Convex Neural Networks}\label{prelimiaries}

We leverage convex neural networks to efficiently learn the complex nonlinear transformations between a highly sparse set of ECG leads. Unlike their non-convex counterparts, convex neural networks are uniquely explainable both intuitively through geometric analysis of the role of optimal hidden neurons, and theoretically through rigorous mathematical proofs underpinning global optimality and the equivalence between convex and non-convex formulations.

The architecture of a two-layer ReLU network was shown to be equivalent to a finite-dimensional second-order cone program (SOCP) \cite{boyd_convex_2004}, which makes it possible to obtain globally optimal parameters using standard convex optimization solvers \cite{pilanci2020neural}. Further work on the properties of convex neural networks have shown that the procedure of globally optimizing deep ReLU networks can be formulated as a Lasso problem \cite{pilanci2023complexity}. For brevity, we summarize the most relevant results here and refer the reader to \cite{pilanci2023complexity} for a complete characterization.

\subsection{The Convex Formulation}
Consider the problem of minimizing the two-layer ReLU network:
\begin{align}\label{nn}
    \min_{\substack{W^{(1)}, W^{(2)}, \\ b^{(1)}, b^{(2)}}} & \ell \left( \sum_{j=1}^m \sigma(XW_j^{(1)} + 1_n b_j^{(1)})W_j^{(2)} + b^{(2)}, y \right) \nonumber \\
    & + \lambda \sum_{j=1}^m \left( ||W_j^{(1)}||_p^2 + ||W_j^{(2)}||_p^2 \right),
\end{align}

where $W^{(1)} \in \mathbb{R}^{d\times m}$, $W^{(2)} \in \mathbb{R}^{m \times c}$ are the weights, $b^{(1)} \in \mathbb{R}^m$, $b^{(2)} \in \mathbb{R}$ are the biases, $\ell$ is a convex loss function, $X \in \mathbb{R}^{n \times d}$ is the training data matrix, $y \in \mathbb{R}^n$ is a vector of labels, and $\lambda > 0$ is a regularization parameter. Here, $\sigma(t)=(t)_{+}=\max(0,t)$ is the ReLU activation and $1_n$ is a length $n$ vector of ones.

In this paper, we consider the special case where $d=1$ and $p=1$. Theorem 1 in \cite{pilanci2023complexity} states that the two-layer neural network \eqref{nn} is equivalent to the following convex $l_1$-regularized problem:
\begin{align}\label{convex_formulation}
    \min_{z \in \mathbb{R}^{2n}, \text{ } t \in \mathbb{R}} \ell(Kz + 1_nt, y) + \lambda ||z||_1.
\end{align}
In \eqref{convex_formulation},  the matrix $K \in \mathbb{R}^{n \times 2n}$ is defined as
\begin{align}\label{K_mat}
    K_{ij} \triangleq \begin{cases}
        (x_i - x_j)_+ & 1 \leq j \leq n \\
        (x_{j-n} - x_i)_+ & n < j \leq 2n,
    \end{cases}
\end{align}
where $x_{i}$ represents the training data points for $1 \leq j \leq n$.
Given the optimal solution $(z^*, t^*)$, an optimal two-layer ReLU network can be constructed as 
\begin{align}\label{function}
    f(x) = \sum_{j=1}^{n} z_j^* (x - x_j)_+ + \sum_{j=1}^{n} z_{j+n}^*(x_j - x)_+ + t^*.
\end{align}
\textit{Remark 1.} Theorem 4 in \cite{pilanci2023complexity} extends the convex formulation to arbitrary dimensions. In dimension $d$, the $K$ matrix is defined as $K_{ij} = \kappa(x_i, x_{j_1},...,x_{j_{d-1}})$, with
\begin{align}\label{eq:K_d_dim}
    \kappa(x, u_1,...,u_{d-1}) &= \frac{(x \wedge u_1 \wedge \cdots \wedge u_{d-1})_+}{||u_1 \wedge \cdots \wedge u_{d-1})||_1}  \notag \\
    &=\frac{\textbf{Vol}_+(\mathcal{P}(x, u_1,...,u_{d-1}))}{||u_1 \wedge \cdots \wedge u_{d-1})||_1}.
\end{align}
In $K_{ij}$, $j = (j_1,...,j_{d-1})$ denotes the indexes over all combinations of $d-1$ rows $x_{j_1},...,x_{j_{d-1}}$ of the data matrix $X$. The symbol $\wedge$ stands for the wedge product and
$\mathcal{P}(x, u_1,...,u_{d-1})$ denotes the parallellotope formed by the vectors $x, u_1,...,u_{d-1}$. The numerator is therefore given by the signed volume of this parallellotope. \hfill $\diamond$

\subsection{Model Interpretability}\label{interpretability}
Each row of \eqref{K_mat} represents a ReLU function with breakpoints at certain training data points. This formulation shows that a globally optimal network can be constructed as a piece-wise linear function using a sum of ReLUs which have their breaklines at a certain subset of data points. In other words, the neurons of an optimal two-layer ReLU network are orthogonal to specific subsets within the training dataset. This insight is critical for model interpretability, as it identifies data subsets which are directly related to model parameters. This offers a distinct advantage over non-convex neural networks where such an insight is not readily available. 

\textit{Remark 2.} In $d$ dimensions, the optimal neurons are orthogonal to $d-1$ points in the training set. Therefore, the hidden neurons are activated on a halfspace defined by the hyperplane that passes through these data points. Further insights can be derived by interpreting the neuron activations as the directional distance to the affine hull of the special subsets of datapoints at the ReLU breaklines. We refer to \cite{pilanci2023complexity} for an in-depth characterization of this geometric interpretation. \hfill $\diamond$

\textit{Remark 3.} The $L_1$ regularization of $z$ corresponds to weight decay, which induces sparsity in the solution, the degree of which is controlled by $\lambda$. Varying $\lambda$ therefore equates to a search over optimal sparse network architectures which results in lightweight models. \hfill $\diamond$

\begin{figure}[!t]

\begin{minipage}[b]{1.00\linewidth}
  \centering
  \centerline{\includegraphics[width=8cm]{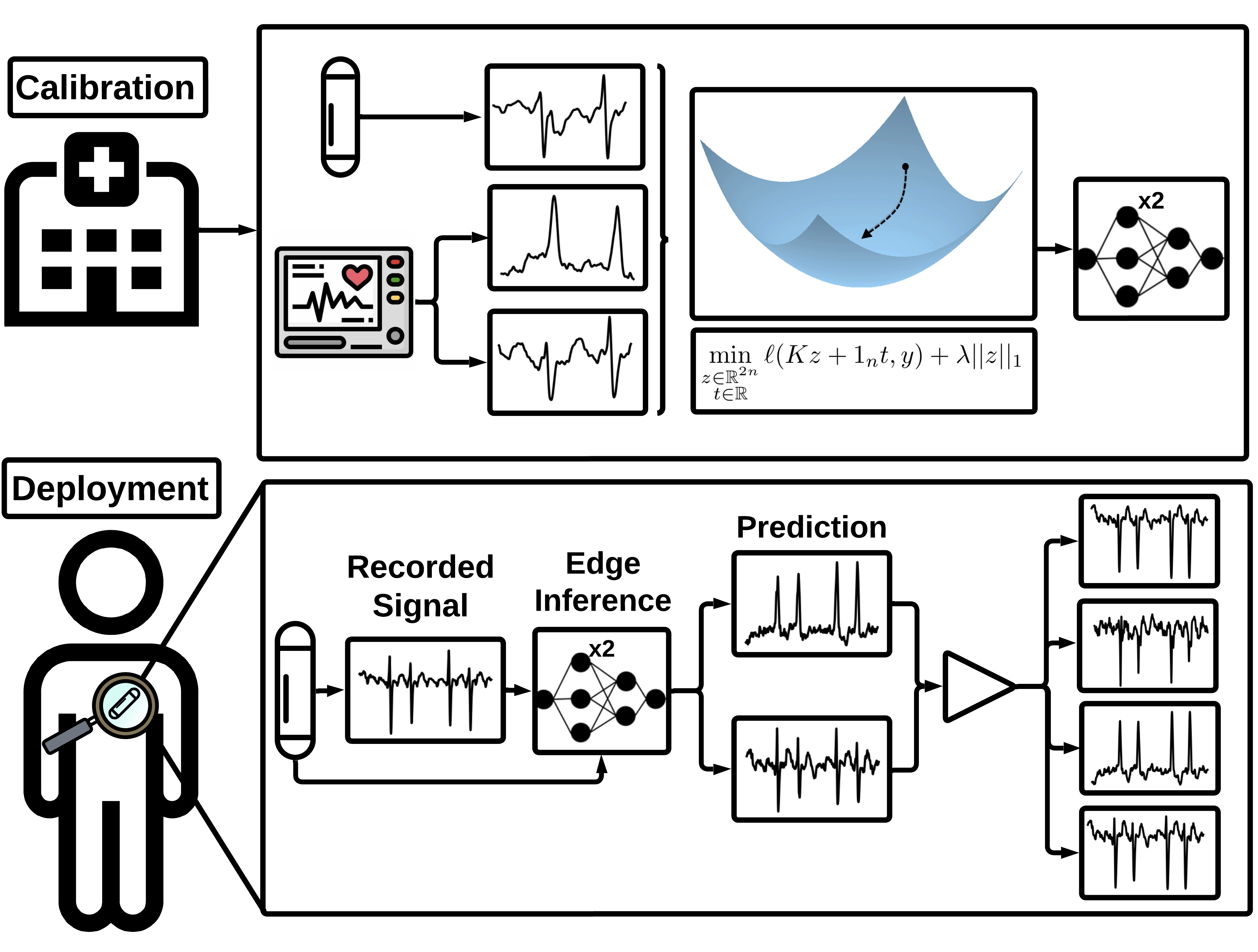}}
\end{minipage}
\caption{Visualization of a Potential Multi-Model Reconstruction Paradigm. In the calibration stage, Leads I and II are recorded along with the ICM device signal to train the coupled ConvexECG model setup. In the deployment stage, the trained models operate on the ICM signal to reconstruct the 6-Lead ECG.}
\label{fig:approach}
\end{figure}

\section{Problem Formulation}\label{problem}

We consider the task of reconstructing a set of 6-lead ECG signals of a patient from a single ICM recording. Specifically, we aim to recover Lead I, Lead II, Lead III, aVR, aVL, and aVF. Let $x^{\text{ICM}} \in \mathbb{R}^n$ represent an ECG recording of length $n$, and let $x^{\text{I}}, x^{\text{II}}, x^{\text{III}}, x^{\text{aVR}}, x^{\text{aVL}}, x^{\text{aVF}}$ denote the reconstruction objectives.

Leads I and II, as described by Einthoven's Triangle, serve as basis vectors spanning the "frontal" plane of a patient's cardiac electrical activity \cite{dupre_basic_2005, ghista_frontal_2010}. By applying linear transformations to these two leads, we can derive Lead III and the augmented limb leads (aVR, aVL, and aVF), thus fully reconstructing the six-lead system.

\textit{Remark 4.} While the theoretical underpinnings are straightforward, the initial mapping from the ICM lead to Leads I and II requires a non-linear model, such as a neural network. In particular, since the ICM is typically placed near the "horizontal" plane, no direct linear transformation exists to the limb leads. \hfill $\diamond$

Let $x^{\text{ICM}}_t$ denote the voltage magnitude of an ICM signal at time $t$. We seek to find transformations $f_{\text{I}}, f_{\text{II}}: \mathbb{R} \rightarrow \mathbb{R}$ that map $x^{\text{ICM}}_t$ to the corresponding, time-aligned values $x^{\text{I}}_t$ and $x^{\text{II}}_t$ in Leads I and II. The remaining frontal plane leads are then derived via Einthoven's principle \cite{dupre_basic_2005}.

\section{Approach}\label{approach}

In this section, we present our method for reduced lead set ECG reconstruction using convex neural networks. Figure \ref{fig:approach} illustrates a potential clinical application where our method could enable continuous remote monitoring of patients. Although the paper does not directly explore the integration of an ICM device, this paradigm envisions a "calibration phase" at the clinic, where Leads I and II are monitored and used to train the neural network. After calibration, now in the "deployment stage," the model operates on ICM signals to remotely reconstruct the 6-Lead ECG. This framing highlights the future possibilities of our approach, while this current work demonstrates the method's efficacy in a simulated setting for reconstructing the 6-Lead ECG.

\subsection{Dataset Construction}
We construct our dataset as follows:
\begin{enumerate}
    \item Given an ICM recording $x^{\text{ICM}}$, we use the first $T$ datapoints $\{x^{\text{ICM}}_t\}_{t=1}^T$ as the input to our convex model.

    \item Given recordings $x^{\text{I}}$ and $x^{\text{II}}$ of Lead I and Lead II, we construct two sets of time-aligned labels $\{x^{\text{I}}_t\}_{t=1}^T$ and $\{x^{\text{II}}_t\}_{t=1}^T$.

    \item Accordingly, we construct two training datasets as
    \begin{align}
        &X^{\text{I}} \triangleq (\{x^{\text{ICM}}_t\}_{t=1}^T, \{x^{\text{I}}_t\}_{t=1}^T), \notag \\
        &X^{\text{II}} \triangleq (\{x^{\text{ICM}}_t\}_{t=1}^T, \{x^{\text{II}}_t\}_{t=1}^T).
    \end{align}
\end{enumerate}

\subsection{Data Preprocessing} Each ECG recording was filtered using a Butterworth filter with frequency cutoffs between 0.5 Hz and 150 Hz to remove baseline drift and high-frequency noise. The signals were then standardized lead by lead to have zero mean and unit variance, ensuring comparability across all leads and reducing potential bias during the reconstruction process. The ECG recordings, originally sampled at 500 Hz (yielding 2500 samples per 5-second window), were then downsampled to 1250 samples to reduce computational load and processing time without significantly affecting signal quality. Segments of length 125 were used for training in all cases, while the remaining 1125 samples were reserved for testing. All experiments were implemented in Python using the MOSEK solver for convex optimization in CVXPY \cite{diamond2016cvxpy, agrawal2018rewriting, MOSEK_Aps_2024}. 

\subsection{Baseline Models} We compare ConvexECG to three baseline models: \begin{enumerate} \item Linear Regression: A simple baseline for lead reconstruction. \item Multi-Layer Perceptron (MLP): A 4-layer ReLU-activated network, optimized using Adam with a fixed learning rate of 0.01. \item Long Short-Term Memory (LSTM): An LSTM network, where hyperparameters (hidden size, layers, learning rate) were optimized using Optuna \cite{akiba2019optuna}. The search space included up to 250 neurons, up to 3 layers, and learning rates between 1e-4 and 1e-1. We ran 100 trials using the Adam optimizer. \end{enumerate} ConvexECG, in this experiment, was set to a constant $L_1$ regularization parameter of $\lambda = 0.01$. The performance of each model was evaluated using the Pearson Correlation between the actual and predicted ECG signals across all patients. Results are presented as the mean performance in the 25 patients.

\subsection{ECG Reconstruction Using Convex Neural Networks}
We model $f_{\text{I}}$ and $f_{\text{II}}$ using the convex formulation of 2-layer ReLU networks defined in section \ref{prelimiaries}. Each neural network is trained using the datasets $X^{\text{I}}$ and $X^{\text{II}}$ respectively to find the governing transformations between the ICM lead and leads I and II. The convexity of our formulation allows us to efficiently obtain the optimal neural networks using any standard off-the-shelf optimization solver. 

After training, the full set of 6 leads can be reconstructed by deriving the remaining leads using the following formulae:
\begin{align}
    &x^{\text{III}} = x^{\text{II}} - x^{\text{I}}, \hspace{1cm} x^{\text{aVR}} = -\frac{1}{2}(x^{\text{I}} + x^{\text{II}}) \notag \\
    &x^{\text{aVL}} = \frac{1}{2}(x^{\text{I}} -  x^{\text{III}}), \hspace{0.3cm} 
 x^{\text{aVF}} = \frac{1}{2}(x^{\text{II}} + x^{\text{III}})
\end{align}

\begin{figure}[htb]

\begin{minipage}[b]{1.05\linewidth}
  \centering
  \centerline{\includegraphics[width=8cm]{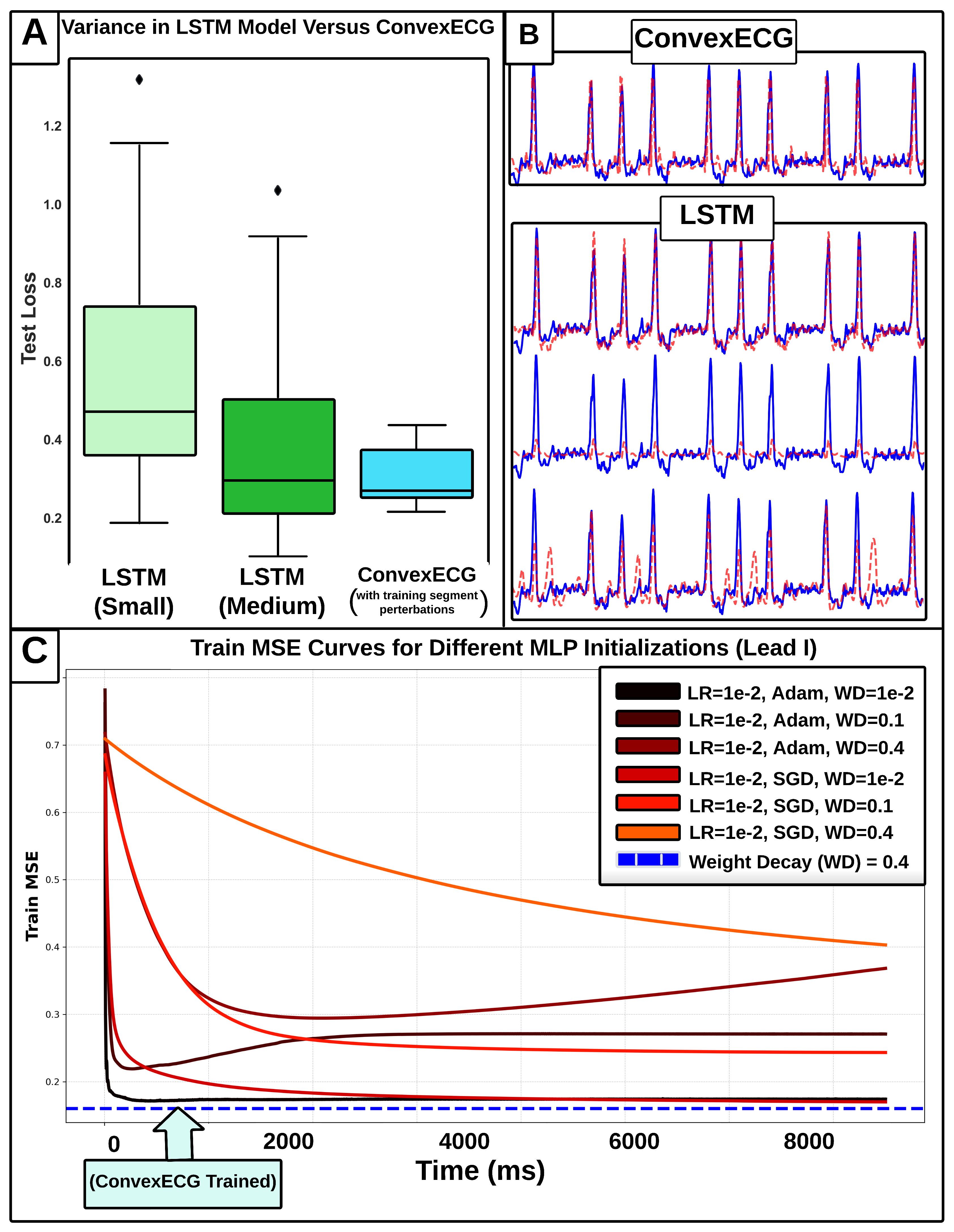}}
\end{minipage}
\caption{Comparison of model variance and performance. \textbf{(A)} Variance in predictions for ConvexECG vs. 2-layer (small) and 3-layer (medium) LSTMs. \textbf{(B)} Test set reconstructions for ConvexECG and LSTM initializations. \textbf{(C)} Train Mean Squared Error curves for 2-layer ReLU MLPs compared to ConvexECG.}
\label{fig:variance}
\end{figure}

\section{Experimental Results}\label{experiments}
In this section, we present the experimental results for evaluating the performance of ConvexECG against the baseline models. The experiments were conducted on ECG data from 25 patients, collected at Stanford Hospital. Waiver of consent form was approved by Stanford Institutional Review Board (IRB) due to the retrospective nature of the study. We compare our approach against linear regression, a multi-layer perceptron (MLP), and a long short-term memory (LSTM) network.

\subsection{Reconstruction Results} 

\begin{figure*}[htb]
\centering
\includegraphics[width=\textwidth]{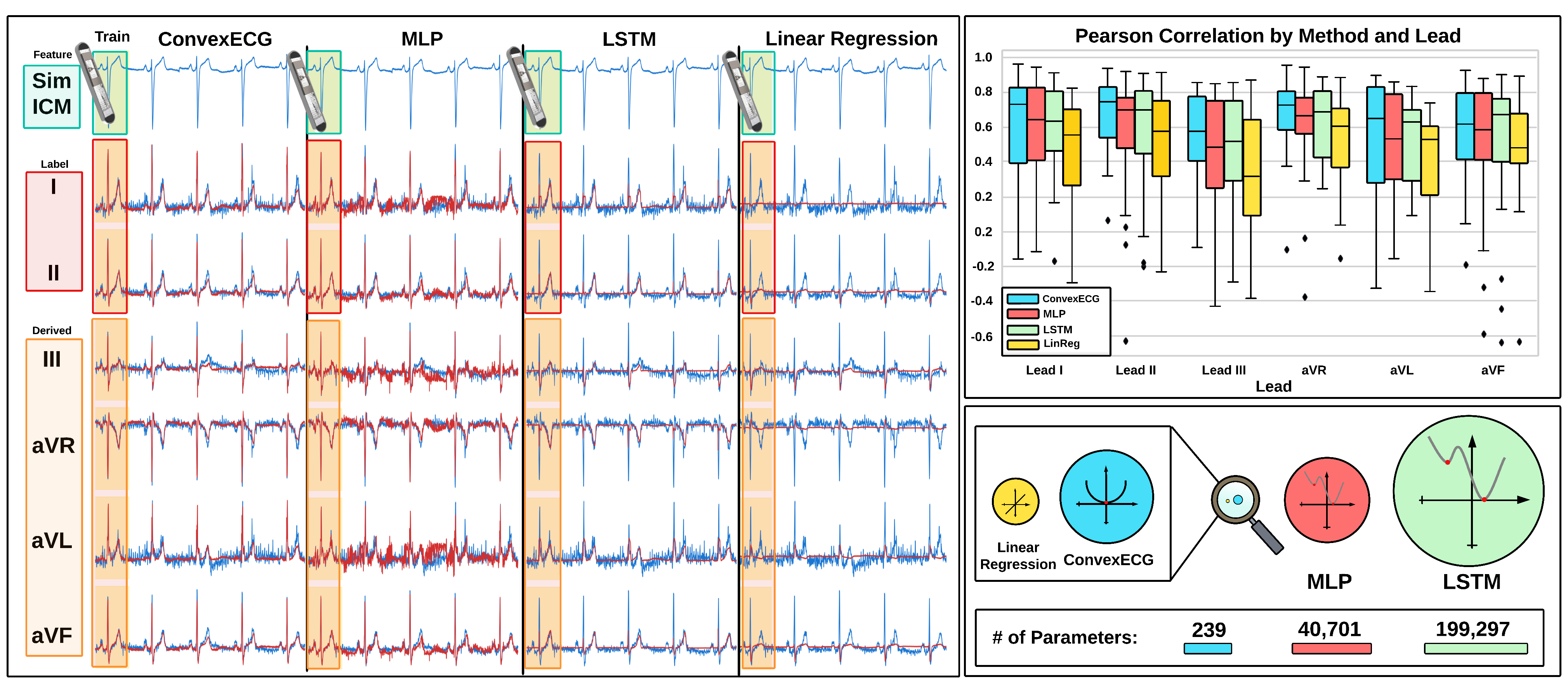}
\caption{Main panel shows 6-lead ECG reconstructions from the simulated ICM signal, with the training segment marked. Top right inset presents a box plot of Pearson correlation coefficients comparing reconstruction quality, while the bottom right inset illustrates model complexity by parameter count.}
\label{fig:res}
\end{figure*}

Figure \ref{fig:res} provides a qualitative comparison of ConvexECG's performance against the baseline models. ConvexECG effectively reconstructs the 6-lead ECG from the ICM input, capturing both linear and non-linear inter-lead dynamics. In contrast, linear regression struggles with more complex inter-lead relationships, often producing overly simplistic predictions, particularly near the isoelectric line. While the non-convex MLP captures the general signal morphology, it is prone to overfitting due to its larger architecture. ConvexECG, with its sparsity and guaranteed global optimality, mitigates this issue, producing cleaner, more accurate reconstructions. The LSTM model performs similarly to ConvexECG but comes with a significantly larger model size, highlighting ConvexECG’s ability to achieve competitive performance with a more resource-efficient and interpretable framework. Additionally, ConvexECG achieves the highest average Pearson correlation across almost all leads, and does so with lower computational cost and greater explainability than the baseline non-convex models.

\subsection{Variance and Convergence Analysis} 
Figure \ref{fig:variance} serves to highlight the robustness of ConvexECG in comparison to non-convex methods (LSTM and MLP). We visualize the detrimental effects of convergence in local minimas for ECG reconstruction by randomly initializing the non-convex methods throughout different training runs and by varying the training hyperparameters. Figure \ref{fig:variance}A illustrates the variance of the test loss of the non-convex models when their parameters are randomly initialized on a fixed dataset. Since the training of ConvexECG is deterministic, we instead perturb train segments via temporal shifts to introduce randomness in the models. Even with the perturbed dataset, ConvexECG yields a lower average test loss and variance compared to the LSTM and MLP. Figure \ref{fig:variance}B visualizes the effects of local minimas on model predictions. Different initializations of the LSTM on the same dataset can result in drastically different training outcomes, leading to a complete inability to capture the transformations in extreme cases. Figure \ref{fig:variance}C further illustrates model variance using the training curves produced by different hyperparemeter configurations. Non-convexity hinders the MLP in reaching the global optimal objective value, which ConvexECG deterministically attains every time. 

\subsection{Analysis of Model Explainability}
\begin{figure}[!t]

\begin{minipage}[b]{1\linewidth}
  \centering
  \centerline{\includegraphics[width=8.5cm]{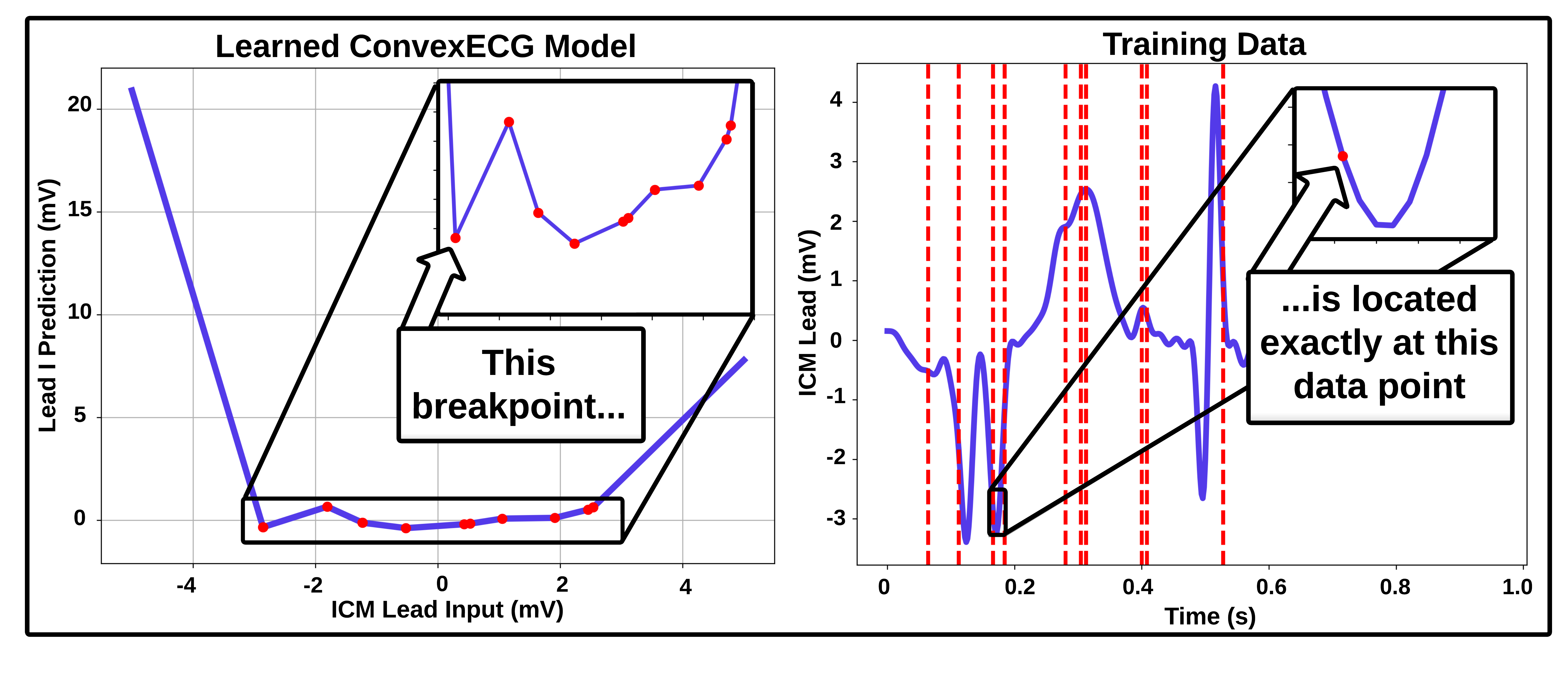}}
\end{minipage}
\caption{The learned $f_{\text{I}}$ plotted against input ICM values along with its training data, illustrating the link between the model’s behavior to specific datapoints.}
\label{fig:interpret}
\end{figure}

Following \cite{amann2020explainability}, we define model explainability as the ability to reconstruct the steps resulting in the model's predictions. To this end, we illustrate the explainability of ConvexECG using figure \ref{fig:interpret}, which visualizes the special subset of points in the training dataset mentioned in Section \ref{interpretability}. From the theory of convex neural networks, we know that the piece-wise linear function \eqref{function} constitutes an optimal two-layer ReLU network, constructed as a sum of ReLU functions with breaklines located at certain subset of datapoints in the training set. This subset therefore defines the basis upon which the model derives its predictions from. The left panel of Figure \ref{fig:interpret} illustrates an example of a function mapping learned by ConvexECG. The breakpoints are easily identified as the set of non-differentiable points (mV values) of the function, shown as red dots in the figure. The right panel illustrates the same points in the training data time-series, found by matching the input values at the breaklines in the left figure to the points in the dataset bearing the same value. ConvexECG therefore enables the re-tracing of its behavior back to the exact subset of training samples which the model used to generate the prediction with. Geometrically, the neuron activations induced by an input can be understood as the oriented distance of that input to the affine hull of this subset of datapoints. In short, ConvexECG is explainable both in terms \textbf{what} it learns (the piece-wise linear function \eqref{function}), \textbf{where} it learns the information from (the special training subset), and \textbf{how} it predicts using the learned information (the geometric interpretation).

\section{Conclusions and Future Work}\label{conclusion}

We presented ConvexECG, a novel method for reconstructing a comprehensive set of six ECG leads from a single ICM lead. Our approach leverages the theory of convex neural networks to model the complex nonlinear inter-lead transformations. We demonstrate our method's effectiveness in reconstructing the full six-lead ECG signals from a single ICM recording while being lightweight and explainable. This paper represents an early effort in adapting convex neural networks to the medical domain with a focus on continuous monitoring of cardiac health. It remains an open research topic to investigate the potential application of more complex convexified network architectures to the task of ECG reconstruction.



\clearpage
\vfill\pagebreak

\label{sec:refs}

\bibliographystyle{IEEEbib}
\bibliography{refs}

\end{document}